\documentclass[sigconf,screen]{acmart}

\AtBeginDocument{%
  }

\setcopyright{acmlicensed}
\copyrightyear{2026}
\acmYear{2026}
\acmDOI{XXXXXXX.XXXXXXX}
\acmConference[Conference acronym 'XX]{Make sure to enter the correct
  conference title from your rights confirmation email}{June 03--05,
  2026}{Woodstock, NY}
\acmISBN{978-1-4503-XXXX-X/2026/06}

\usepackage{multirow}
\usepackage{arydshln}

\setcopyright{none}
\renewcommand\footnotetextcopyrightpermission[1]{}

\begin{document}

\title{When 2D Cues Fail: Improving Image Manipulation Localization with Reliable 3D Geometry}


\author{Guofeng Yu}
\email{yuguofeng@stu.xju.edu.cn}
\affiliation{
  \institution{School of Computer Science and Technology, Xinjiang University}
  \city{Urumqi}
  \country{China}
}

\author{Zhiqing Guo}
\authornote{Corresponding author.}
\email{guozhiqing@xju.edu.cn}
\affiliation{
  \institution{School of Computer Science and Technology, Xinjiang University}
  \city{Urumqi}
  \country{China}
}

\author{Dan Ma}
\email{madan@xju.edu.cn}
\affiliation{
  \institution{School of Computer Science and Technology, Xinjiang University}
  \city{Urumqi}
  \country{China}
}

\author{Gaobo Yang}
\email{yanggaobo@hnu.edu.cn}
\affiliation{
  \institution{College of Computer Science and Electronic Engineering}
  \city{Hunan University, Changsha}
  \country{China}
}

\renewcommand{\shortauthors}{Trovato et al.}

\begin{abstract}

Existing image manipulation localization (IML) methods rely heavily on 2D forensic cues, such as low-level artifacts, noise traces, and semantic inconsistencies in the manipulated image. While effective in many cases, these cues become much less discriminative when manipulated regions are well blended with their surrounding context in appearance. In such cases, a manipulated region may remain locally appearance-consistent, but still violate the geometric structure of the surrounding scene. This limitation motivates us to go beyond purely 2D evidence and introduce geometric reasoning into IML. To this end, we leverage monocular reconstruction to obtain auxiliary geometric cues, including depth and surface normals. However, a key challenge lies in the fact that reconstructed geometry on manipulated images is inherently noisy and cannot be used naively. Rather than treating depth and normals as direct evidence, we estimate their reliability and exploit them selectively for localization. Based on this principle, we design a geometry-aware framework (GFrame) that fuses reliable geometric cues with RGB features and propagates them across scales to improve fine-grained localization. Extensive experiments show that the proposed method achieves excellent performance under limited budget constraints. These results indicate that reliable 3D geometry provides complementary forensic evidence beyond traditional 2D cues for IML. Related code will be released.

\end{abstract}






\maketitle
\pagestyle{empty}

\section{Introduction}

\begin{figure}[t]
    \centering
    \includegraphics[width=\columnwidth]{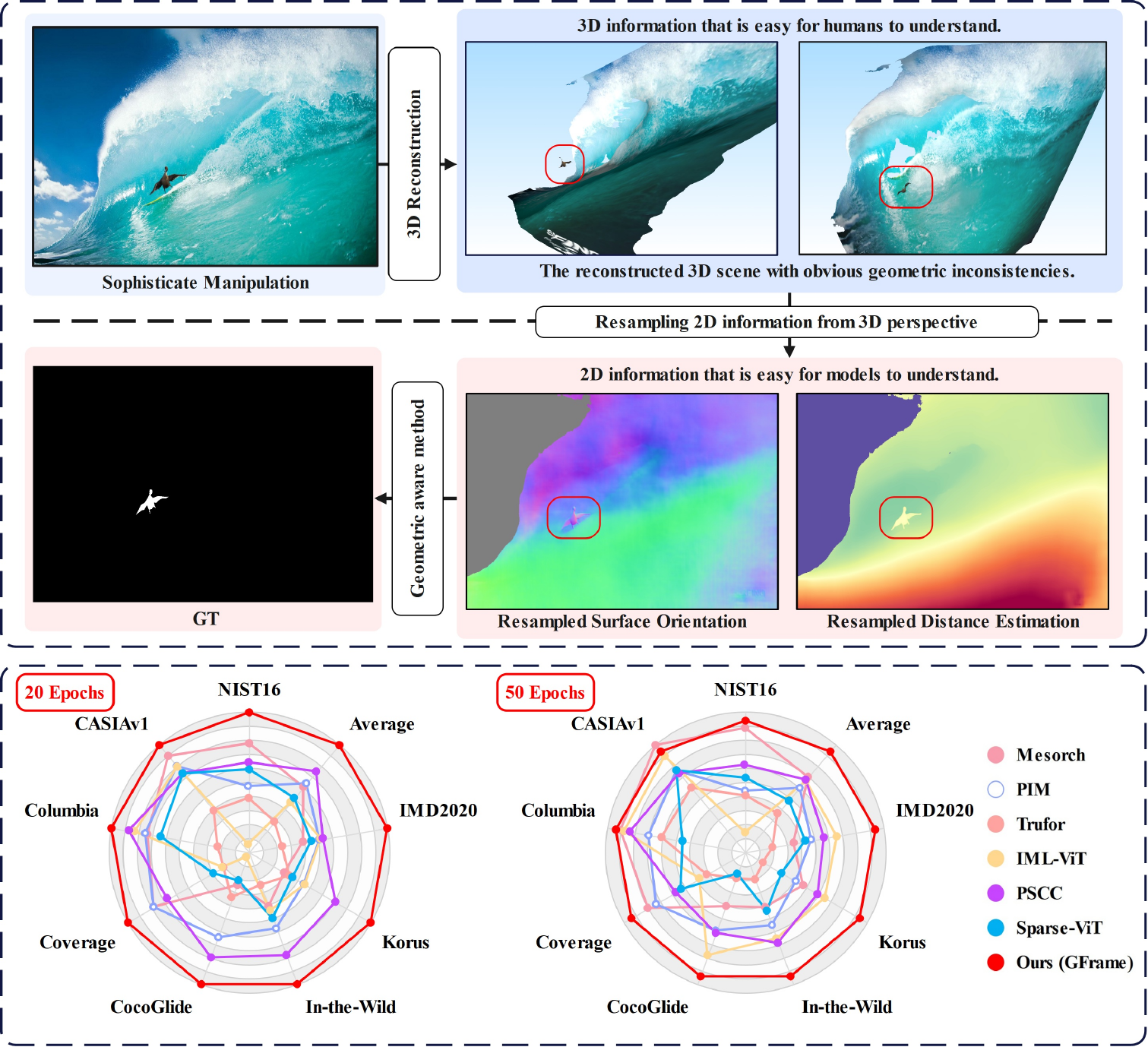}
    \caption{Even when manipulated regions are visually well blended, they may still violate the underlying scene geometry. Reconstructed depth and surface normals can reveal such geometric inconsistencies and provide complementary cues for image manipulation localization.}
    \label{fig:firstpic}
\end{figure}

Image manipulation localization (IML) aims to identify and precisely localize manipulated regions within an image. It serves as a fundamental research problem in the field of multimedia forensics. With the rapid development of image editing tools and deep generative models, diffusion-based and GAN-based generation techniques have been widely applied\cite{Karras2019stylegan2,rombach2022high}. Manipulated content has become increasingly realistic and can hardly be detected only from visual appearance. Therefore, designing robust localization models against high-quality manipulations is of great significance to ensure the credibility of digital media.


Existing IML methods mainly fall into two categories. The first category relies on explicit physical or statistical clues, such as CFA inconsistencies, JPEG artifacts, lens distortion, lighting, or shadow constraints, to expose tampered regions~\cite{bianchi2012image,ferrara2012image,fu2012forgery,gloe2010efficient}. While interpretable, these methods often depend on strong assumptions and handcrafted modeling, which limits their applicability in complex real-world conditions. The second category is data-driven 2D localization, where CNN-based~\cite{dong2022mvss,guillaro2023trufor,wu2019mantra,hu2020span} and transformer-based~\cite{ma2023iml,su2025can,zhu2025mesoscopic} models learn to detect pixel-level anomalies, noise traces, and semantic inconsistencies directly from manipulated images. These methods have significantly improved benchmark performance, but they remain largely appearance-driven and may struggle when manipulations preserve local texture continuity and semantic plausibility. Meanwhile, recent advances in monocular 3D reconstruction have made it increasingly practical to recover geometric cues such as depth and surface normals from a single image~\cite{wang2025moge,wang2025moge2,yin2023metric3d}. However, such reconstructed geometry has not been systematically explored for IML. In particular, current IML methods still lack a practical framework for exploiting geometric cues when depth and normal estimation is noisy and inaccurate.

This limitation points to a largely unexplored opportunity for IML. Even when a manipulated region is visually well integrated into its surrounding context, it may still violate the underlying geometric structure of the scene. Such violations may manifest as implausible depth discontinuities, inconsistent surface orientation, or local geometry that conflicts with the global scene layout. These inconsistencies are often difficult to infer from appearance alone, but may become accessible through reconstructed depth and surface normals. However, directly introducing reconstructed geometry into IML is far from trivial. Monocular reconstruction on manipulated images is inherently imperfect, and the resulting geometric estimates may suffer from ambiguity, local artifacts, or outright reconstruction failure. Therefore, the central challenge is not simply to incorporate geometry, but to exploit reconstructed geometric cues in a selective and stable manner, so that useful structural evidence can complement strong RGB representations without being overwhelmed by reconstruction noise.

To address this challenge, we propose GFrame, a geometry-aware framework for IML. GFrame reconstructs depth and surface normals from the input image and uses them as complementary structural cues alongside RGB features. To cope with the noise and ambiguity of reconstructed geometry, we introduce an Uncertainty-aware Geometric Fusion Module (UGFM), which uses confidence-inspired reliability cues to regulate how geometric features are injected into the RGB stream. We further design a Cross-Scale Query Module (CSQM) to propagate scene-level geometric context across resolutions, thereby improving the interaction between global structural reasoning and fine-grained boundary localization. In this way, reconstructed geometry is not treated as direct forensic evidence, but as selectively exploited auxiliary information for robust manipulation localization. Our main contributions are as follows:
\begin{itemize}
    \item We introduce a geometry-aware formulation for image manipulation localization that leverages reconstructed depth and surface normals as complementary structural cues beyond conventional 2D forensic evidence.
    \item We propose GFrame, which combines reliability-aware geometric fusion with cross-scale structural propagation, enabling the model to selectively exploit noisy reconstructed geometry for fine-grained localization.
    \item We conduct experiments on both in-distribution and out-of-distribution benchmarks, showing that geometry-aware cues provide a competitive and promising complement to strong 2D IML baselines.
\end{itemize}

\section{Related Work}

\subsection{Traditional Methods Based on Physical Cues}


Early IML methods mainly relied on explicit modeling of physical or statistical inconsistencies introduced during manipulation. Representative studies exploited clues from the imaging pipeline, such as JPEG compression artifacts\cite{bianchi2012image}, CFA inconsistencies\cite{ferrara2012image}, and lens distortion\cite{fu2012forgery}, as well as physical constraints including lighting direction and shadow relationships~\cite{gloe2010efficient}. These methods are interpretable and effective when their underlying assumptions hold. However, they heavily depend on handcrafted modeling and are sensitive to post-processing or real-world degradations, which restricts their robustness in unconstrained forensic scenarios. In contrast, our work does not rely on manually specified physical rules, but explores reconstructed geometry as a complementary structural signal within a learnable manipulation localization framework.


\subsection{Deep Learning-based IML Methods}

With the rise of deep learning, IML has largely shifted from handcrafted analysis to data-driven feature learning. CNN-based methods such as MVSS-Net\cite{dong2022mvss}, TruFor\cite{guillaro2023trufor}, Mantra-Net\cite{wu2019mantra}, and SPAN\cite{hu2020span} focus on detecting local anomalies, noise traces, or pixel-level inconsistencies from manipulated images. More recent transformer-based and hybrid architectures like IML-ViT\cite{ma2023iml}, Sparse-ViT\cite{su2025can} and Mesorch~\cite{zhu2025mesoscopic} further strengthen long-range modeling and semantic reasoning, improving localization accuracy on benchmark datasets. Despite their strong empirical performance, these methods remain fundamentally appearance-driven: they infer manipulations primarily from RGB evidence and may still struggle when manipulated regions are visually well integrated into the surrounding context. Our work differs in that it investigates whether reconstructed geometric cues can provide complementary evidence beyond conventional 2D appearance features.


\subsection{Geometry Priors and 3D Reconstruction}


Geometry priors have proved valuable in a wide range of vision tasks, where depth and surface orientation provide structural information that is difficult to infer from appearance alone~\cite{wang2025moge,wang2025moge2,yin2023metric3d}. Recent advances in monocular 3D reconstruction have further made it highly practical to robustly recover geometric cues such as depth and surface normals from a single image, enabling geometry-aware reasoning without requiring multi-view input or explicit 3D supervision~\cite{Zhang2019PAP,ranftl2021vision,Wang2023DaCOD}. However, the use of reconstructed geometry for IML remains largely underexplored. In particular, geometry reconstructed from manipulated images is inherently noisy and spatially unreliable, which makes direct use of such cues non-trivial. Our work addresses this gap by investigating how to selectively exploit reconstructed depth and normals through reliability-aware fusion and cross-scale propagation for robust manipulation localization.

\begin{figure*}
    \centering
    \includegraphics[width=\linewidth]{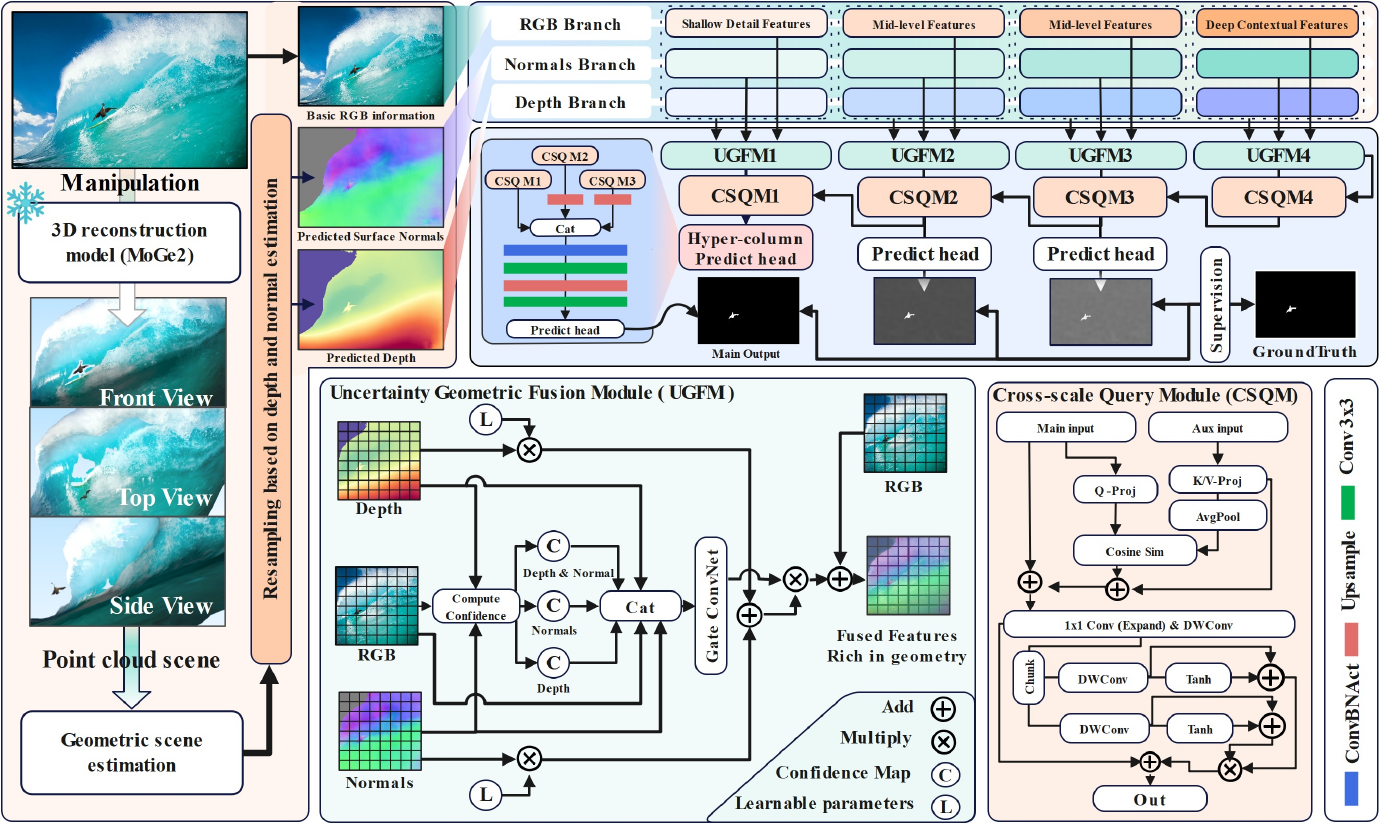}
    \caption{Overview of GFrame. Given an input RGB image, we first reconstruct depth and surface normal maps as auxiliary geometric cues. Multi-scale RGB, depth, and normal features are then fused by UGFM through reliability-aware gating, which selectively injects trustworthy geometry into the RGB stream. CSQM further propagates coarse structural context to finer resolutions in a top-down manner, improving the interaction between global geometric reasoning and fine-grained localization.}
    \label{fig:model_main}
\end{figure*}

\section{Methodology}

In this paper, we propose GFrame, a geometry-aware framework for image manipulation localization. The core idea is to complement conventional RGB evidence with reconstructed geometric cues while explicitly accounting for their unreliability. Given an input RGB image, we first reconstruct two auxiliary geometric maps, namely depth and surface normals, which provide scene-structure information beyond appearance. However, because these cues are estimated from images that may have already been manipulated, they are inherently noisy and cannot be uniformly trusted across the entire image. To address this, GFrame is designed around a simple principle: reconstructed geometry should be exploited selectively rather than being uniformly injected into the localization pipeline.


\subsection{Geometric Fusion of UGFM}

GFrame starts from three aligned inputs: the RGB image, the reconstructed depth map, and the reconstructed surface normal map. The latter two are obtained directly from monocular reconstruction and serve as explicit geometric cues beyond appearance. Intuitively, depth describes the spatial layout of the scene, while surface normals characterize local surface orientation. If reconstructed reliably, these cues can expose structural inconsistencies that are difficult to infer from RGB evidence alone. However, because the reconstruction is performed on already manipulated images, the resulting depth and normal maps are not uniformly trustworthy: some regions preserve meaningful scene geometry, whereas others may be affected by ambiguity, local artifacts, or reconstruction failure. The role of UGFM is therefore to regulate how reconstructed geometry is injected into the RGB stream according to its local reliability.

For each encoder stage $i$, let $F^i_{rgb}$, $F^i_d$, and $F^i_n$ denote the RGB, depth, and normal features, respectively. To determine whether the reconstructed geometry is sufficiently stable to assist localization, we derive a set of confidence-inspired geometric cues from the reconstructed maps themselves. We begin with the depth map $D$. A locally reliable depth estimate is expected to vary smoothly except at genuine structural transitions, whereas unstable reconstruction often manifests as irregular second-order fluctuations. Based on this observation, we define a depth confidence term:
\begin{equation}
C_d = \exp(-|\nabla^2 D|),
\end{equation}
where $\nabla^2 D$ denotes the Laplacian of the reconstructed depth. This term assigns higher confidence to regions whose depth variation is locally coherent.

We then turn to the reconstructed normal map $N_p$. Surface normals are expected to be unit vectors by construction, so deviations from unit magnitude often indicate unstable reconstruction or poorly formed local geometry. We therefore define a normal confidence term:
\begin{equation}
C_n = \exp\left(-\left|\|N_p\|_2 - 1\right|\right),
\end{equation}
which down-weights regions where the normal estimate is internally less reliable.

Depth and normals should also be mutually compatible, since they describe the same underlying scene geometry from different perspectives. To capture this compatibility, we derive a normal field $N_d$ from the reconstructed depth and compare it with the reconstructed normals:
\begin{equation}
A_{dn} = \frac{1}{2}(N_p \cdot N_d + 1).
\end{equation}
A high value of $A_{dn}$ indicates that the two geometric sources are locally consistent, whereas a low value suggests that at least one of them is unreliable. Importantly, $C_d$, $C_n$, and $A_{dn}$ are not intended as exact uncertainty estimates. Instead, they act as practical reliability cues that help identify where reconstructed geometry is likely to be informative for localization.

Using these cues, we form the fusion input and a spatial gate:
\begin{equation}
F^i_{in} = [F^i_{rgb}, F^i_d, F^i_n, C^i_d, C^i_n, A^i_{dn}],
\end{equation}
\begin{equation}
G^i = \sigma(\mathrm{Conv}_{gate}(F^i_{in})),
\end{equation}
where $\sigma(\cdot)$ is the sigmoid function. The stage-wise fused representation is then computed as:
\begin{equation}
Z^i = F^i_{rgb} + G^i \odot \left(w_d F^i_d + w_n F^i_n\right),
\end{equation}
where $w_d$ and $w_n$ are learnable scalar weights.

This formulation reflects the intended role of reconstructed geometry in our framework. RGB remains the primary representation because it provides the most stable evidence for localization, while depth and normals serve as complementary structural signals. The residual form preserves the original RGB stream, and the gate allows geometric information to contribute only where its local reliability is sufficiently supported. In this way, UGFM encourages meaningful reconstructed geometry to strengthen localization while suppressing unstable geometric responses before they interfere with appearance-based feature learning.

\subsection{Cross-Scale Query of CSQM}

After geometric fusion, the model must still address another challenge: geometric information is often most informative at coarse scales, whereas manipulation localization ultimately requires precise boundary delineation at high spatial resolution. A useful decoder should therefore not only aggregate multi-scale features, but also propagate scene-level structural context to fine-resolution representations in a targeted manner. This motivates our cross-scale query design.

At decoder stage $i$, a finer-scale feature serves as the query input $Q_{in}$, while a coarser-scale feature serves as the key-value input $KV_{in}$. The coarser feature provides broader structural context, while the finer feature retains local spatial detail. Instead of treating the two scales symmetrically, we explicitly let the fine feature query the coarse feature, because the localization task is fundamentally about refining high-resolution predictions under scene-level geometric constraints.

We compute cross-scale attention using cosine similarity:
\begin{equation}
\mathrm{Attn} =
\mathrm{softmax}
\left(
\frac{Q_{ctx}(K'_{ctx})^\top}{\|Q_{ctx}\| \|K'_{ctx}\|}\cdot \tau
\right)V'_{ctx},
\end{equation}
where $Q_{ctx}$, $K'_{ctx}$, and $V'_{ctx}$ denote the transformed query, key, and value features, respectively, and $\tau$ is a learnable temperature parameter. We adopt cosine attention because the features being matched come from different resolutions and may differ in magnitude; cosine similarity emphasizes directional agreement rather than absolute scale, leading to more stable cross-scale matching.

The attended representation is first added back to the query feature through a residual connection, producing an intermediate feature $Q'$. To further enhance local detail while preserving the global structural guidance obtained from cross-scale attention, we introduce a gated feed-forward block based on a depthwise-convolutional inverted bottleneck.

Specifically, $Q'$ is first normalized and expanded along the channel dimension, and the expanded tensor is then split into two branches, denoted by $X_1$ and $X_2$. Each branch is processed by a lightweight depthwise-convolutional transformation with a \texttt{Tanh} nonlinearity and a residual connection, yielding transformed features $X_1'$ and $X_2'$. The refined output is then computed as:
\begin{equation}
P_{out} = Q' + \mathrm{Conv}_{1\times1}^{\mathrm{proj}}(X_1' \odot X_2'),
\end{equation}
where $\odot$ denotes element-wise multiplication.

This design complements cross-scale attention in two ways. First, depthwise convolutions introduce spatially aware local context at substantially lower cost than full convolutions. Second, the split-and-multiply structure provides pixel-wise, data-dependent modulation, allowing one branch to act as a content path and the other as a dynamic gate. Compared with a standard MLP-style feed-forward block that only mixes channels independently at each location, this design is better suited for emphasizing or suppressing local manipulation artifacts after global structural interaction.

The output $P_{out}$ serves as the refined feature at the current scale. It is then upsampled and passed to the next finer stage as the key-value input, forming a top-down refinement cascade. In this way, scene-level geometric context is progressively propagated to high-resolution representations, improving the interaction between global structural reasoning and precise manipulation boundaries.

\begin{table*}[!t]
\centering
\renewcommand{\arraystretch}{1.1}
\setlength{\tabcolsep}{3.5pt}
\caption{
Main comparison on InD and OOD benchmarks using F1@0.5.
Results are reported under 20 and 50 epochs when available.
Best and second-best results are marked in \textbf{bold} and \underline{underline}, respectively.
}
\begin{tabular*}{\textwidth}{@{\extracolsep{\fill}}l c cccc @{\hspace{5mm}} cccccc}
\toprule
\multirow{2}{*}{\textbf{Method}} &
\multirow{2}{*}{\textbf{Epoch}} &
\multicolumn{4}{c}{\textbf{In-distribution (F1@0.5)}} &
\multicolumn{6}{c}{\textbf{Out-of-distribution (F1@0.5)}} \\
\cmidrule(lr){3-6} \cmidrule(lr){7-12}
& & CASIAv1 & Coverage & NIST16 & \textbf{Avg.}
& Columbia & In-the-Wild & Korus & CocoGlide & IMD2020 & \textbf{Avg.} \\
\midrule
PSCC        & 20 & 0.460 & 0.398 & 0.357 & 0.405 & 0.690 & \underline{0.354} & \underline{0.289} & \underline{0.303} & 0.287 & 0.289 \\
(TCSVT'22)  & 50 & 0.489 & 0.361 & 0.367 & 0.406 & 0.710 & 0.333 & 0.187 & 0.249 & 0.325 & 0.274 \\
\cdashline{1-12}
TruFor      & 20 & 0.240 & 0.126 & 0.214 & 0.193 & 0.180 & 0.113 & 0.100 & 0.133 & 0.128 & 0.118 \\
(CVPR'23)   & 50 & 0.398 & 0.200 & 0.241 & 0.280 & 0.516 & 0.101 & 0.047 & 0.079 & 0.113 & 0.085 \\
\cdashline{1-12}
IML-ViT     & 20 & 0.495 & 0.130 & 0.030 & 0.218 & 0.657 & 0.201 & 0.137 & 0.017 & 0.279 & 0.158 \\
(AAAI'24)   & 50 & 0.599 & 0.238 & 0.083 & 0.307 & 0.747 & 0.314 & 0.207 & 0.318 & \underline{0.371} & \underline{0.303} \\
\cdashline{1-12}
Mesorch     & 20 & 0.560 & 0.465 & 0.433 & 0.486 & 0.584 & 0.183 & 0.087 & 0.094 & 0.211 & 0.144 \\
(AAAI'25)   & 50 & \textbf{0.667} & \underline{0.505} & \underline{0.524} & \underline{0.565} & \underline{0.767} & 0.201 & 0.153 & 0.165 & 0.201 & 0.180 \\
\cdashline{1-12}
Sparse-ViT  & 20 & 0.462 & 0.176 & 0.330 & 0.323 & 0.511 & 0.230 & 0.107 & 0.083 & 0.239 & 0.164 \\
(AAAI'25)   & 50 & 0.505 & 0.333 & 0.316 & 0.385 & 0.384 & 0.211 & 0.095 & 0.068 & 0.249 & 0.156 \\
\cdashline{1-12}
PIM         & 20 & 0.505 & 0.464 & 0.260 & 0.409 & 0.596 & 0.264 & 0.134 & 0.245 & 0.272 & 0.229 \\
(TPAMI'25)  & 50 & -- & -- & -- & -- & -- & -- & -- & -- & -- & -- \\
\midrule
GFrame      & 20 & \underline{0.627} & \textbf{0.592} & \textbf{0.556} & \textbf{0.592} & \textbf{0.797} & \textbf{0.450} & \textbf{0.296} & \textbf{0.381} & \textbf{0.528} & \textbf{0.490} \\
(Ours)      & 50*& 0.610 & 0.670 & 0.583 & 0.621 & 0.808 & 0.424 & 0.273 & 0.388 & 0.518 & 0.482 \\
\bottomrule
\end{tabular*}

\label{tab:main_results}
\end{table*}

\subsection{Boundary-Aware Optimization}

The most informative geometric discrepancies in manipulated images often concentrate around structural transitions, object contours, and foreground-background interfaces. This observation suggests that the optimization objective should emphasize boundary-sensitive learning rather than treating all pixels equally. Otherwise, the contribution of geometry-aware cues may be diluted by large homogeneous regions that are easier to classify but less informative.

We therefore construct a boundary-aware structural objective. Given the ground-truth mask $M$, we define a spatial weighting map:
\begin{equation}
W = 1 + \lambda \left| \mathrm{AvgPool}_k(M) - M \right|,
\end{equation}
which assigns larger weights to pixels near mask transitions. Based on $W$, we define a structural loss:
\begin{equation}
L_{\mathrm{struct}} = L_{\mathrm{wBCE}}(Y,M,W) + L_{\mathrm{wIoU}}(Y,M,W),
\end{equation}
where $Y$ denotes the predicted manipulation mask. The weighted binary cross-entropy term improves pixel-level discrimination, while the weighted IoU term promotes region-level consistency under boundary emphasis.

To further encourage clear separation between manipulated and authentic regions, we introduce a margin-based regularization term:
\begin{equation}
L_{\mathrm{margin}} = \frac{1}{N}\sum_p \max(0, m - S_p),
\end{equation}
where $S_p$ denotes the signed prediction confidence at pixel $p$. The role of this term is not to replace the segmentation objective, but to discourage uncertain predictions around the decision boundary. In our setting, this is particularly useful because the geometric cues introduced by the model are often most informative near structural discontinuities; encouraging sharper prediction separation helps the network absorb such cues more effectively during training.

The final objective combines the boundary-aware structural term, the margin regularization term, and auxiliary supervision from intermediate predictions at lower resolutions. The auxiliary losses are assigned normal weights in practice and mainly serve to stabilize optimization. Overall, this objective is designed to strengthen boundary learning and improve how geometry-aware cues are translated into localization performance.

\begin{table}[!t]
\centering
\caption{The dataset used in our experiments. CM, SP, and IP indicate three common image manipulation types: copy-move, splicing, and inpainting.}
\resizebox{0.47\textwidth}{!}{%
\begin{tabular}{@{}ccccccc@{}}
\toprule
\textbf{Dataset} & \textbf{Nums} & \textbf{\#CM} & \textbf{\#SP} & \textbf{\#IP} & \textbf{Train} & \textbf{Test} \\ \midrule
CASIAv2~\cite{dong2013casia}        & 5123  &3295   &1828   &0      & 5123  & 0   \\ 
Coverage~\cite{wen2016coverage}     & 100   &100    &  0    &0      & 70    & 30  \\ 
NIST16~\cite{guan2019mfc}           & 564   &68     &288    &208    & 383   & 181 \\ 
CASIAv1~\cite{dong2013casia}        & 920   &459    &461    &0      & 0     & 920 \\
Columbia~\cite{hsu2006columbia}     & 180   &0      &180    &0      & 0     & 180 \\ 
CocoGlide~\cite{nichol2021glide}    & 512   & -     & -     & -     & 0     & 512 \\
In-the-Wild ~\cite{huh2018fighting} & 201   & 0     & 201   & 0     & 0     & 201 \\
Korus~\cite{korus2016evaluation}    & 220   & -     &-      &-      & 0     & 220 \\
IMD2020~\cite{Novozamsky_2020_WACV} & 2010 &-&-&-&0&2010\\
\bottomrule
\end{tabular}
}
\label{tab:datasets}
\end{table}

\section{Experiment}
\subsection{Datasets and Metrics}
We evaluate GFrame on eight widely-used IML datasets as shown in Table.\ref{tab:datasets}. We divide them into two groups: in-distribution (InD) datasets used for training and validation, and out-of-distribution (OOD) datasets used only for generalization evaluation. All models are trained on the InD training sets and evaluated on both the InD test sets and the OOD test sets. The exact dataset assignment and split protocol are provided in the supplementary material for reproducibility. We note that the NIST16 data used in our experiments are manually cleaned to avoid data leakage.

We use the F1-score as the primary evaluation metric, following common practice in IML benchmarks. As recent work IMDLBenCo\cite{ma2024imdl} has pointed out that AUC may overestimate performance. We therefore report F1@0.5 as the main metric, which provides a stricter and more interpretable evaluation of IML masks.

\subsection{Implementation Details}
We implement GFrame in PyTorch 2.1.0 (Python 3.10, CUDA 12.1) and train on a Tesla T4 GPU. The model utilizes PvtV2-B2 as the RGB backbone and PvtV2-B1 for the geometric branch, both pre-trained on ImageNet. All inputs are resized to 512x512. We follow the normalization and data augmentation strategies from IMDLBenco\cite{ma2024imdl}. We compare with six methods: PSCC\cite{liu2022pscc}, Trufor\cite{guillaro2023trufor}, IML-ViT\cite{ma2023iml}, Mesorch\cite{zhu2025mesoscopic}, Sparse-ViT\cite{su2025can}, and PIM\cite{kong2025pixel}, all re-evaluated under the IMDLBenco framework for fair comparison.

\begin{table*}[!t]
\centering
\renewcommand{\arraystretch}{1.1}
\setlength{\tabcolsep}{3.5pt}
\caption{Ablation study of GFrame on both InD and OOD benchmarks. In addition to component-level removals, we also compare different geometric input configurations, including depth-only (RGB-D), normal-only (RGB-N), and different reconstruction method (Metric3D). The results show that geometric cues are consistently useful with UGFM, CSQM and the loss.}
\begin{tabular*}{\textwidth}{@{\extracolsep{\fill}}lcccc@{\hspace{6mm}}cccccc}
\toprule
\multirow{2}{*}{\textbf{Settings}} &
\multicolumn{4}{c}{\textbf{In-distribution (F1@0.5)}} &
\multicolumn{6}{c}{\textbf{Out-of-distribution (F1@0.5)}} \\
\cmidrule(lr){2-5} \cmidrule(lr){6-11}
& CASIAv1 & Coverage & NIST16 & \textbf{Avg.}
& Columbia & In-the-Wild & Korus & CocoGlide & IMD2020 & \textbf{Avg.} \\
\midrule
GFrame         & 0.627 & 0.592 & 0.556 & 0.592          & 0.797 & 0.450 & 0.296 & 0.381 & 0.528 & 0.490 \\
Metric3D       & 0.614 & 0.568 & 0.538 & 0.573          & 0.767 & 0.444 & 0.307 & 0.371 & 0.523 & 0.482 \\
RGB-D          & 0.603 & 0.564 & 0.529 & 0.566          & 0.788 & 0.493 & 0.321 & 0.355 & 0.540 & 0.500 \\
RGB-N          & 0.606 & 0.548 & 0.554 & 0.569          & 0.800 & 0.444 & 0.305 & 0.400 & 0.529 & 0.495 \\
Backbone       & 0.505 & 0.589 & 0.563 & 0.552          & 0.736 & 0.488 & 0.299 & 0.282 & 0.526 & 0.466 \\
\cdashline{1-11}
w/o UGFM       & 0.594 & 0.553 & 0.545 & 0.564          & 0.765 & 0.406 & 0.291 & 0.336 & 0.501 & 0.460 \\
w/o CSQM       & 0.574 & 0.513 & 0.571 & 0.553          & 0.743 & 0.398 & 0.286 & 0.356 & 0.498 & 0.456 \\
w/o Geo.branch & 0.567 & 0.548 & 0.545 & 0.553          & 0.742 & 0.421 & 0.267 & 0.309 & 0.505 & 0.449 \\
w/o Loss       & 0.563 & 0.539 & 0.490 & 0.531          & 0.699 & 0.315 & 0.157 & 0.249 & 0.419 & 0.368 \\
\bottomrule
\end{tabular*}
\label{tab:ablation}
\end{table*}

\subsection{Main Results}

Table~\ref{tab:main_results} reports the main quantitative comparison on both InD and OOD benchmarks. Under the matched 20-epoch training budget, GFrame outperforms all compared methods on both settings. We further train GFrame for 50 epochs to examine the effect of longer optimization. Compared with the 20-epoch checkpoint, the 50-epoch model further improves the InD average F1 from 0.592 to 0.621, while the OOD average changes from 0.490 to 0.482.This indicates that the proposed geometry-aware design remains effective under both limited and extended training budgets, with the 20-epoch checkpoint providing the best cross-distribution trade-off.

On the InD benchmarks, GFrame-20 already achieves an average F1 of 0.592, surpassing all previous methods, including the strongest reported 50-epoch baseline, Mesorch, which reaches 0.565. With longer training, GFrame-50 further raises the InD average to 0.621 and achieves the best scores on Coverage and NIST16, while remaining competitive on CASIAv1. These results suggest that reliability-aware geometric cues improve not only early-stage training effectiveness, but also the in-domain performance ceiling under extended optimization.

On the OOD benchmarks, GFrame shows consistently strong cross-dataset generalization. GFrame-20 achieves the best OOD average of 0.490, and GFrame-50 still remains the strongest among all compared methods with an OOD average of 0.482. The slight decrease from 20 to 50 epochs suggests that, although longer optimization benefits in-domain fitting, the best generalization trade-off is reached earlier. Interestingly, a similar non-monotonic trend can also be observed in several appearance-driven baselines, whose OOD performance does not consistently improve with longer training. Overall, these results support our claim that reliable geometric cues provide transferable structural evidence beyond conventional 2D appearance features, especially under distribution shift.

\subsection{Ablation Studies}

We conduct ablation studies to evaluate the main components of GFrame. The results are reported in Table~\ref{tab:ablation}, including three geometry-input variants (RGB-D, RGB-N, and Metric3D) and four component ablations (w/o UGFM, CSQM, Geo.branch, and Loss).

Introducing geometric cues consistently improves performance over the RGB-only variant. Removing the geometric branch reduces the average F1 to 0.553 on InD and 0.449 on OOD, confirming that reconstructed geometry provides useful complementary evidence. RGB-D and RGB-N further show that both depth and normal cues are individually effective, while replacing the reconstruction source with Metric3D leads to a slight drop, suggesting that localization quality depends on the reliability of the reconstructed geometry.

\begin{figure}[!t]
    \centering
    \includegraphics[width=\linewidth]{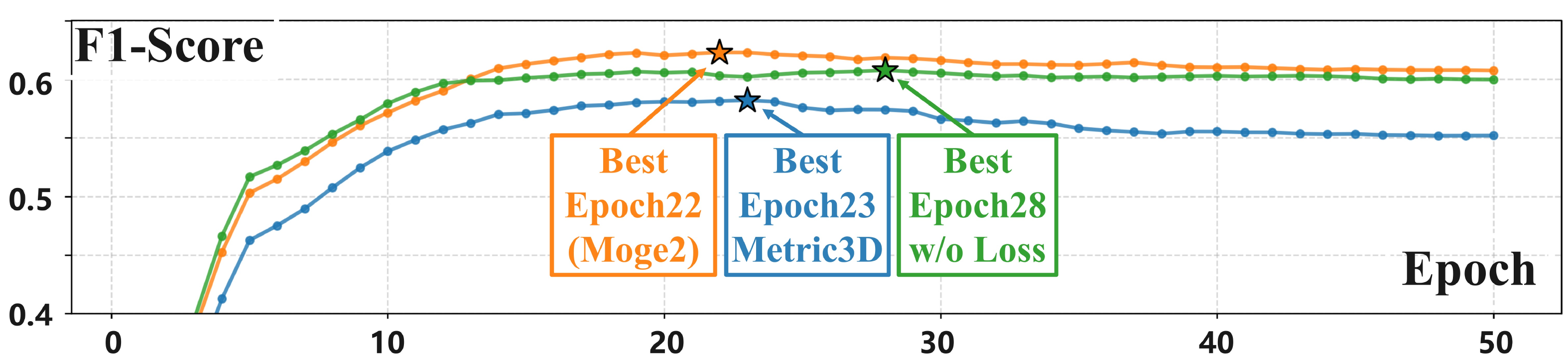}
    \caption{Validation F1 curves of representative settings. The proposed training loss leads to earlier attainment of peak performance than the counterpart without the loss term.}
    \label{fig:loss_curve}
\end{figure}

Removing UGFM or CSQM also degrades performance, indicating that the gain comes not only from geometry itself, but also from how it is fused and propagated. Without the proposed loss, performance drops most significantly, to 0.531 on InD and 0.368 on OOD, showing that boundary-aware optimization is important for fully exploiting geometry-aware cues.

\begin{table*}[!t]
\centering
\small
\setlength{\tabcolsep}{5.7mm}
\renewcommand{\arraystretch}{1.1}
\caption{Robustness analysis under standard perturbations on CASIAv1 using F1@0.5. We evaluate Gaussian noise, Gaussian blur, and JPEG compression at increasing perturbation levels. GFrame achieves the best average performance across all settings.}
\begin{tabular}{ll|ccccccc}
\hline
\hline
\multirow{2}{*}{\textbf{Method}} &
\multicolumn{7}{c}{\textbf{GaussNoise (Standard Deviations)}} &
\multirow{2}{*}{\textbf{Avg.}} \\
\cline{2-8}
& None & 3 & 7 & 11 & 15 & 19 & 23 \\
\hline
IML-ViT    & 0.495 & 0.456 & 0.425 & 0.354 & 0.289 & 0.237 & 0.200 & 0.326 \\
Mesorch    & 0.560 & 0.525 & 0.463 & 0.369 & 0.319 & 0.276 & 0.254 & 0.368 \\
Sparse-ViT & 0.462 & 0.414 & 0.304 & 0.165 & 0.104 & 0.071 & 0.067 & 0.187 \\
PIM        & 0.505 & 0.465 & 0.385 & 0.284 & 0.237 & 0.195 & 0.161 & 0.286 \\
Ours       & 0.627 & 0.583 & 0.574 & 0.574 & 0.553 & 0.554 & 0.548 & 0.564 \\
\hline
\hline

\multirow{2}{*}{\textbf{Method}} &
\multicolumn{7}{c}{\textbf{GaussianBlur (Kernel Size)}} &
\multirow{2}{*}{\textbf{Avg.}} \\
\cline{2-8}
& None & 3 & 7 & 11 & 15 & 19 & 23 \\
\hline
IML-ViT    & 0.495 & 0.374 & 0.265 & 0.137 & 0.039 & 0.007 & 0.001 & 0.130 \\
Mesorch    & 0.560 & 0.440 & 0.318 & 0.158 & 0.064 & 0.029 & 0.016 & 0.171 \\
Sparse-ViT & 0.462 & 0.402 & 0.308 & 0.156 & 0.074 & 0.040 & 0.043 & 0.171 \\
PIM        & 0.505 & 0.349 & 0.237 & 0.113 & 0.034 & 0.015 & 0.007 & 0.126 \\
Ours       & 0.627 & 0.537 & 0.466 & 0.366 & 0.276 & 0.207 & 0.179 & 0.338 \\
\hline
\hline

\multirow{2}{*}{\textbf{Method}} &
\multicolumn{7}{c}{\textbf{JpegCompression (Quality Factors)}} &
\multirow{2}{*}{\textbf{Avg.}} \\
\cline{2-8}
& None & 100 & 90 & 80 & 70 & 60 & 50 \\
\hline
IML-ViT    & 0.495 & 0.466 & 0.443 & 0.347 & 0.327 & 0.322 & 0.284 & 0.370 \\
Mesorch    & 0.560 & 0.544 & 0.489 & 0.305 & 0.305 & 0.255 & 0.185 & 0.347 \\
Sparse-ViT & 0.462 & 0.434 & 0.355 & 0.182 & 0.153 & 0.109 & 0.089 & 0.219 \\
PIM        & 0.505 & 0.503 & 0.435 & 0.267 & 0.212 & 0.142 & 0.076 & 0.306 \\
Ours       & 0.627 & 0.537 & 0.490 & 0.359 & 0.333 & 0.287 & 0.225 & 0.372 \\
\hline
\end{tabular}
\label{tab:robust}
\end{table*}

\begin{table}[t]
\setlength{\tabcolsep}{1.2mm}
\renewcommand{\arraystretch}{1.2}
\centering
\caption{Performance on images processed by social media. GFrame outperforms baselines in these real-world scenarios.}
\begin{tabular}{lccccc} 
\toprule
\multirow{2}{*}{\textbf{Method}} &
\multicolumn{5}{c}{\textbf{Online Social Media Compression (F1-score)}}\\
\cline{2-6} 
& Facebook & WeChat & Weibo & WhatsApp & \textbf{Avg.} \\
\midrule 
SparseViT & 0.388 & 0.244 & 0.410 & 0.389 & 0.358 \\
PIM & 0.438 & 0.308 & 0.465 & 0.463 & 0.419 \\
IML-ViT & 0.468 & 0.343 & 0.482 & 0.465 & 0.440 \\
Mesorch & 0.499 & 0.364 & 0.514 & 0.510 & 0.472 \\
\cdashline{1-6}
Ours      & 0.569 & 0.459 & 0.575 & 0.581 & 0.546 \\
\bottomrule 
\end{tabular}

\label{tab:osn}
\end{table}

This effect is further supported by the validation F1 curves in Figure.\ref{fig:loss_curve}. As shown in the figure, the full objective reaches its best validation performance earlier than the counterpart without the loss term. In particular, the default MoGe2-based configuration peaks around epoch 22, whereas the version without the loss reaches its best result later, around epoch 28. A similar early-peak behavior can also be observed for the Metric3D-based variant. These observations suggest that the proposed objective not only improves the final localization quality, but also facilitates earlier absorption of boundary-sensitive structural cues during training.

\subsection{Robustness Analysis}

Table~\ref{tab:robust} reports the robustness results on CASIAv1 under Gaussian noise, Gaussian blur, and JPEG compression. GFrame achieves the best average performance across all three perturbation types. The advantage is more pronounced under noise and blur, suggesting that the proposed geometry-aware design is less dependent on fragile local appearance statistics when RGB cues are degraded.

We further evaluate images processed by mainstream online social-media platforms in Table~\ref{tab:osn}. GFrame achieves the best performance on all four platforms, including Facebook, WeChat, Weibo, and WhatsApp. These results further support that reliable geometric cues provide stronger cross-domain robustness than purely RGB-driven baselines in realistic scenarios.



\begin{figure*}
    \centering
    \includegraphics[width=\textwidth]{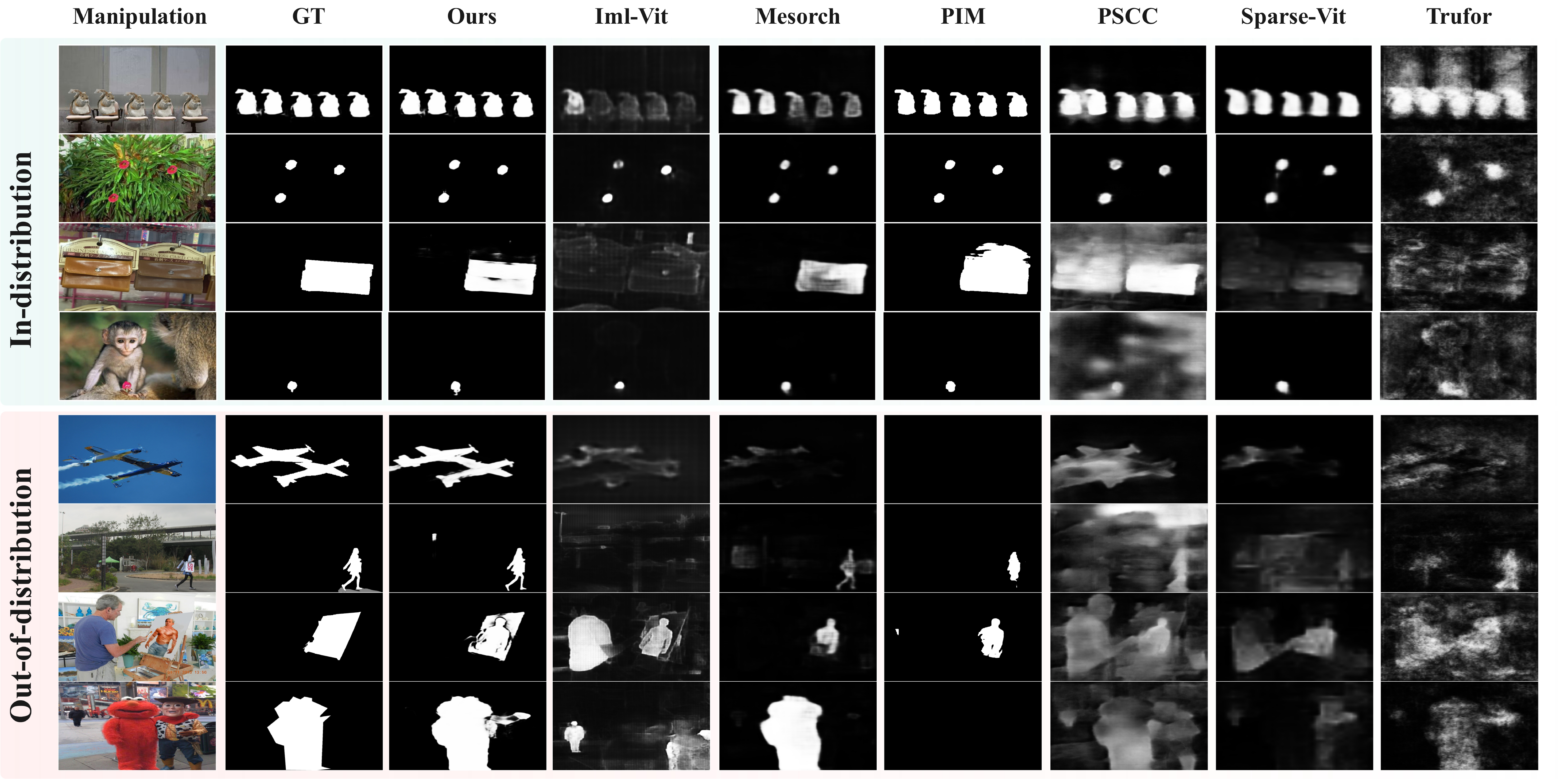}
    \caption{Qualitative comparison on in-distribution (InD) and out-of-distribution (OOD) samples. Each row corresponds to one manipulation case, and each column shows the prediction of a different method, including the ground truth and our result. Compared with representative baselines, GFrame produces more complete manipulated regions, cleaner background suppression, and more accurate boundary localization across diverse scenarios.}
\label{fig:comparison}
\end{figure*}


\begin{table}[t]
\centering
\small
\setlength{\tabcolsep}{2.5mm}
\renewcommand{\arraystretch}{1.1}
\caption{Model complexity comparison. GFrame achieves a favorable balance between performance and cost.}
\begin{tabular}{lccc}
\toprule
\textbf{Method} & \textbf{Size} & \textbf{Params (M)} & \textbf{FLOPs (G)} \\
\midrule
PSCC        & $256 \times 256$   & 3.7   & 45.7  \\
Sparse-ViT  & $512 \times 512$   & 50.3  & 46.2  \\
TruFor      & $512 \times 512$   & 68.7  & 236.5 \\
Mesorch     & $512 \times 512$   & 85.8  & 124.9 \\
IML-ViT     & $1024 \times 1024$ & 91.0  & 448.0 \\
PIM         & $512 \times 512$   & 152.5 & 682.9 \\
\midrule
GFrame      & $512 \times 512$   & 53.44 & 70.56 \\
\midrule
MoGe2       & $512 \times 512$   & 330.90 & 1588.27 \\
Metric3D    & $512 \times 512$   & 37.50  & 56.80 \\
\midrule
GFrame + MoGe2    & $512 \times 512$ & 384.34 & 1658.83 \\
GFrame + Metric3D & $512 \times 512$ & 90.94  & 127.36 \\
\bottomrule
\end{tabular}

\label{tab:complexity}
\end{table}

\subsection{Complexity Analysis}
We further compare the model complexity of GFrame with representative baselines in Table.\ref{tab:complexity}. Considering only the main localization network, GFrame contains 53.44M parameters and 70.56G FLOPs at an input resolution of $512 \times 512$. This is substantially lighter than several recent transformer-based or hybrid IML models, indicating that the performance gain of GFrame does not come from an excessively heavy localization architecture.

At the same time, GFrame relies on an external monocular reconstruction model to provide depth and normal cues, and the total deployment cost therefore depends strongly on the selected reconstruction backend. As shown in Table.\ref{tab:complexity}, MoGe2 provides the stronger default configuration in our experiments but introduces a large additional cost, whereas Metric3D is considerably more efficient and yields a much lower total complexity. This observation is consistent with the ablation results, where different reconstruction sources exhibit a clear accuracy--efficiency trade-off.

Overall, the complexity analysis suggests that the proposed geometry-aware localization network itself is relatively efficient, while the practical deployment cost is mainly governed by the reconstruction model. This also indicates that future improvements in lightweight monocular geometry estimation could directly benefit the efficiency of geometry-aware IML.

\subsection{Qualitative Results}
We further present qualitative comparisons in Figure.\ref{fig:comparison} to examine the localization behavior of GFrame on both InD and OOD samples. The figure compares our predictions with several representative baselines under diverse manipulation scenarios.

Overall, GFrame produces prediction masks that are more spatially complete and structurally coherent. On in-distribution samples, our method more accurately recovers manipulated regions with complex shapes or weak local artifacts, while suppressing irrelevant responses in the background. Compared with several RGB-driven baselines, the predictions of GFrame are generally cleaner and better aligned with the ground-truth masks, especially near object contours and fine structural boundaries.

The advantage becomes more evident on OOD samples, where many competing methods exhibit fragmented activation, incomplete foreground recovery, or strong responses to unrelated textures. In contrast, GFrame maintains more stable foreground localization and cleaner background suppression across diverse scenes. This qualitative behavior is consistent with the quantitative results in the previous sections and suggests that the proposed geometry-aware design provides a useful complement to conventional 2D appearance cues, particularly in challenging cross-domain cases.

\section{Limitations and Future Work}
Although GFrame introduces geometric cues into IML, the current framework remains image-based. Its geometric information is reconstructed from a single RGB image, and part of the original structural information may be lost during this conversion process, which in turn limits the effectiveness of downstream localization. In addition, depth and surface normals are still indirect forms of 3D structure and may not fully capture more complex geometric inconsistencies in manipulated content.


A promising direction for future work is to move beyond image-level reconstructed geometry and explore direct 3D representations, such as point clouds, meshes, or implicit scene representations, for manipulation localization. These representations may preserve richer spatial relationships and surface structures than depth and normal maps, helping detect geometric inconsistencies that are difficult to capture in 2D projections. Efficient 3D feature extraction and cross-modal alignment therefore remain important directions for robust multimedia forensics.

\section{Conclusion}
We presented GFrame, a geometry-aware framework for image manipulation localization that leverages reconstructed depth and surface normals as complementary structural cues. By combining reliability-aware geometric fusion, cross-scale structural propagation, and boundary-aware optimization, GFrame achieves strong performance on both in-distribution and out-of-distribution benchmarks. The results suggest that geometric cues provide useful complementary evidence beyond conventional 2D appearance-based features for robust and generalizable manipulation localization. More broadly, this work highlights the potential of incorporating 3D structural reasoning into image forensics.

\bibliographystyle{ACM-Reference-Format}
\bibliography{samples/main}

@article{fu2012forgery,
  title={Forgery authentication in extreme wide-angle lens using distortion cue and fake saliency map},
  author={Fu, Huazhu and Cao, Xiaochun},
  journal={IEEE Transactions on Information Forensics and Security},
  volume={7},
  number={4},
  pages={1301--1314},
  year={2012},
  publisher={IEEE}
}

@inproceedings{gloe2010efficient,
  title={Efficient estimation and large-scale evaluation of lateral chromatic aberration for digital image forensics},
  author={Gloe, Thomas and Borowka, Karsten and Winkler, Antje},
  booktitle={Media Forensics and Security II},
  volume={7541},
  pages={62--74},
  year={2010},
  organization={SPIE}
}

@article{ferrara2012image,
  title={Image forgery localization via fine-grained analysis of CFA artifacts},
  author={Ferrara, Pasquale and Bianchi, Tiziano and De Rosa, Alessia and Piva, Alessandro},
  journal={IEEE Transactions on Information Forensics and Security},
  volume={7},
  number={5},
  pages={1566--1577},
  year={2012},
  publisher={IEEE}
}

@article{bianchi2012image,
  title={Image forgery localization via block-grained analysis of JPEG artifacts},
  author={Bianchi, Tiziano and Piva, Alessandro},
  journal={IEEE Transactions on Information Forensics and Security},
  volume={7},
  number={3},
  pages={1003--1017},
  year={2012},
  publisher={IEEE}
}

@inproceedings{wu2019mantra,
  title={Mantra-net: Manipulation tracing network for detection and localization of image forgeries with anomalous features},
  author={Wu, Yue and AbdAlmageed, Wael and Natarajan, Premkumar},
  booktitle={Proceedings of the IEEE/CVF conference on computer vision and pattern recognition},
  pages={9543--9552},
  year={2019}
}

@inproceedings{hu2020span,
  title={SPAN: Spatial pyramid attention network for image manipulation localization},
  author={Hu, Xuefeng and Zhang, Zhihan and Jiang, Zhenye and Chaudhuri, Syomantak and Yang, Zhenheng and Nevatia, Ram},
  booktitle={European conference on computer vision},
  pages={312--328},
  year={2020},
  organization={Springer}
}

@article{dong2022mvss,
  title={Mvss-net: Multi-view multi-scale supervised networks for image manipulation detection},
  author={Dong, Chengbo and Chen, Xinru and Hu, Ruohan and Cao, Juan and Li, Xirong},
  journal={IEEE Transactions on Pattern Analysis and Machine Intelligence},
  volume={45},
  number={3},
  pages={3539--3553},
  year={2022},
  publisher={IEEE}
}

@inproceedings{guillaro2023trufor,
  title={Trufor: Leveraging all-round clues for trustworthy image forgery detection and localization},
  author={Guillaro, Fabrizio and Cozzolino, Davide and Sud, Avneesh and Dufour, Nicholas and Verdoliva, Luisa},
  booktitle={Proceedings of the IEEE/CVF conference on computer vision and pattern recognition},
  pages={20606--20615},
  year={2023}
}

@article{ma2023iml,
  title={IML-ViT: Benchmarking Image Manipulation Localization by Vision Transformer},
  author={Ma, Xiaochen and Du, Bo and Jiang, Zhuohang and Hammadi, Ahmed Y Al and Zhou, Jizhe},
  journal={arXiv preprint arXiv:2307.14863},
  year={2023}
}

@inproceedings{su2025can,
  title={Can we get rid of handcrafted feature extractors? sparsevit: Nonsemantics-centered, parameter-efficient image manipulation localization through spare-coding transformer},
  author={Su, Lei and Ma, Xiaochen and Zhu, Xuekang and Niu, Chaoqun and Lei, Zeyu and Zhou, Ji-Zhe},
  booktitle={Proceedings of the AAAI Conference on Artificial Intelligence},
  volume={39},
  number={7},
  pages={7024--7032},
  year={2025}
}

@inproceedings{zhu2025mesoscopic,
  title={Mesoscopic insights: orchestrating multi-scale \& hybrid architecture for image manipulation localization},
  author={Zhu, Xuekang and Ma, Xiaochen and Su, Lei and Jiang, Zhuohang and Du, Bo and Wang, Xiwen and Lei, Zeyu and Feng, Wentao and Pun, Chi-Man and Zhou, Ji-Zhe},
  booktitle={Proceedings of the AAAI Conference on Artificial Intelligence},
  volume={39},
  number={10},
  pages={11022--11030},
  year={2025}
}

@article{kong2025pixel,
  title={Pixel-inconsistency modeling for image manipulation localization},
  author={Kong, Chenqi and Luo, Anwei and Wang, Shiqi and Li, Haoliang and Rocha, Anderson and Kot, Alex C},
  journal={IEEE Transactions on Pattern Analysis and Machine Intelligence},
  year={2025},
  publisher={IEEE}
}

@article{liu2022pscc,
  title={PSCC-Net: Progressive spatio-channel correlation network for image manipulation detection and localization},
  author={Liu, Xiaohong and Liu, Yaojie and Chen, Jun and Liu, Xiaoming},
  journal={IEEE Transactions on Circuits and Systems for Video Technology},
  volume={32},
  number={11},
  pages={7505--7517},
  year={2022},
  publisher={IEEE}
}

@inproceedings{Karras2019stylegan2,
  title     = {Analyzing and Improving the Image Quality of {StyleGAN}},
  author    = {Tero Karras and Samuli Laine and Miika Aittala and Janne Hellsten and Jaakko Lehtinen and Timo Aila},
  booktitle = {Proc. CVPR},
  year      = {2020}
}

@inproceedings{Wang2023DaCOD,
  author    = {Qingwei Wang and Jinyu Yang and Xiaosheng Yu and Fangyi Wang and Peng Chen and Feng Zheng},
  title     = {Depth-aided Camouflaged Object Detection},
  booktitle = {Proceedings of the 31st ACM International Conference on Multimedia (ACM MM)},
  year      = {2023},
  pages     = {3297--3306},
  doi       = {10.1145/3581783.3611874}
}

@inproceedings{Zhang2019PAP,
  author    = {Zhenyu Zhang and Zhen Cui and Chunyan Xu and Yan Yan and Nicu Sebe and Jian Yang},
  title     = {Pattern-Affinitive Propagation across Depth, Surface Normal and Semantic Segmentation},
  booktitle = {Proceedings of the IEEE/CVF Conference on Computer Vision and Pattern Recognition (CVPR)},
  year      = {2019},
  pages     = {7144–7153}
}

@inproceedings{wang2025moge,
  title={Moge: Unlocking accurate monocular geometry estimation for open-domain images with optimal training supervision},
  author={Wang, Ruicheng and Xu, Sicheng and Dai, Cassie and Xiang, Jianfeng and Deng, Yu and Tong, Xin and Yang, Jiaolong},
  booktitle={Proceedings of the Computer Vision and Pattern Recognition Conference},
  pages={5261--5271},
  year={2025}
}

@misc{wang2025moge2,
      title={MoGe-2: Accurate Monocular Geometry with Metric Scale and Sharp Details}, 
      author={Ruicheng Wang and Sicheng Xu and Yue Dong and Yu Deng and Jianfeng Xiang and Zelong Lv and Guangzhong Sun and Xin Tong and Jiaolong Yang},
      year={2025},
      eprint={2507.02546},
      archivePrefix={arXiv},
      primaryClass={cs.CV},
      url={https://arxiv.org/abs/2507.02546}, 
}

@inproceedings{yin2023metric3d,
  title={Metric3d: Towards zero-shot metric 3d prediction from a single image},
  author={Yin, Wei and Zhang, Chi and Chen, Hao and Cai, Zhipeng and Yu, Gang and Wang, Kaixuan and Chen, Xiaozhi and Shen, Chunhua},
  booktitle={Proceedings of the IEEE/CVF international conference on computer vision},
  pages={9043--9053},
  year={2023}
}

@article{ma2024imdl,
  title={Imdl-benco: A comprehensive benchmark and codebase for image manipulation detection \& localization},
  author={Ma, Xiaochen and Zhu, Xuekang and Su, Lei and Du, Bo and Jiang, Zhuohang and Tong, Bingkui and Lei, Zeyu and Yang, Xinyu and Pun, Chi-Man and Lv, Jiancheng and others},
  journal={Advances in Neural Information Processing Systems},
  volume={37},
  pages={134591--134613},
  year={2024}
}

@inproceedings{dong2013casia,
  title={Casia image tampering detection evaluation database},
  author={Dong, Jing and Wang, Wei and Tan, Tieniu},
  booktitle={2013 IEEE China summit and international conference on signal and information processing},
  pages={422--426},
  year={2013},
  organization={IEEE}
}

@inproceedings{guan2019mfc,
  title={MFC datasets: Large-scale benchmark datasets for media forensic challenge evaluation},
  author={Guan, Haiying and Kozak, Mark and Robertson, Eric and Lee, Yooyoung and Yates, Amy N and Delgado, Andrew and Zhou, Daniel and Kheyrkhah, Timothee and Smith, Jeff and Fiscus, Jonathan},
  booktitle={2019 IEEE Winter Applications of Computer Vision Workshops (WACVW)},
  pages={63--72},
  year={2019},
  organization={IEEE}
}

@article{hsu2006columbia,
  title={Columbia uncompressed image splicing detection evaluation dataset},
  author={Hsu, J and Chang, SF},
  journal={Columbia DVMM Research Lab},
  volume={6},
  year={2006}
}

@inproceedings{wen2016coverage,
  title={COVERAGE—A novel database for copy-move forgery detection},
  author={Wen, Bihan and Zhu, Ye and Subramanian, Ramanathan and Ng, Tian-Tsong and Shen, Xuanjing and Winkler, Stefan},
  booktitle={2016 IEEE international conference on image processing (ICIP)},
  pages={161--165},
  year={2016},
  organization={IEEE}
}

@INPROCEEDINGS{Novozamsky_2020_WACV,
author = {Novozamsky, Adam and Mahdian, Babak and Saic, Stanislav},
title = {IMD2020: A Large-Scale Annotated Dataset Tailored for Detecting Manipulated Images},
booktitle = {2020 IEEE Winter Applications of Computer Vision Workshops (WACVW)},
year = {2020},
month = {March},
pages = {71-80}
}

@article{nichol2021glide,
  title={Glide: Towards photorealistic image generation and editing with text-guided diffusion models},
  author={Nichol, Alex and Dhariwal, Prafulla and Ramesh, Aditya and Shyam, Pranav and Mishkin, Pamela and McGrew, Bob and Sutskever, Ilya and Chen, Mark},
  journal={arXiv preprint arXiv:2112.10741},
  year={2021}
}

@inproceedings{huh2018fighting,
  title={Fighting fake news: Image splice detection via learned self-consistency},
  author={Huh, Minyoung and Liu, Andrew and Owens, Andrew and Efros, Alexei A},
  booktitle={Proceedings of the European conference on computer vision (ECCV)},
  pages={101--117},
  year={2018}
}

@inproceedings{korus2016evaluation,
  title={Evaluation of random field models in multi-modal unsupervised tampering localization},
  author={Korus, Pawe{\l} and Huang, Jiwu},
  booktitle={2016 IEEE international workshop on information forensics and security (WIFS)},
  pages={1--6},
  year={2016},
  organization={IEEE}
}

@inproceedings{rombach2022high,
  title={High-resolution image synthesis with latent diffusion models},
  author={Rombach, Robin and Blattmann, Andreas and Lorenz, Dominik and Esser, Patrick and Ommer, Bj{\"o}rn},
  booktitle={Proceedings of the IEEE/CVF conference on computer vision and pattern recognition},
  pages={10684--10695},
  year={2022}
}

@inproceedings{ranftl2021vision,
  title={Vision transformers for dense prediction},
  author={Ranftl, Ren{\'e} and Bochkovskiy, Alexey and Koltun, Vladlen},
  booktitle={Proceedings of the IEEE/CVF international conference on computer vision},
  pages={12179--12188},
  year={2021}
}










\end{document}